\documentclass{article}

\PassOptionsToPackage{numbers}{natbib}
\usepackage[final]{neurips_2023}

\usepackage{graphicx}
\usepackage{amsmath}
\usepackage[utf8]{inputenc} % allow utf-8 input
\usepackage[T1]{fontenc}    % use 8-bit T1 fonts
\usepackage{hyperref}       % hyperlinks
\usepackage{url}            % simple URL typesetting
\usepackage{booktabs}       % professional-quality tables
\usepackage{amsfonts}       % blackboard math symbols
\usepackage{nicefrac}       % compact symbols for 1/2, etc.
\usepackage{microtype}      % microtypography
\usepackage{xcolor}         % colors
\usepackage{lmodern}

\title{Detecting Spurious Correlations via Robust Visual Concepts in Real and AI-Generated Image Classification}

\author{%
  Preetam Prabhu Srikar Dammu\\
  University of Washington\\
  Seattle, WA, USA \\
  \texttt{preetams@uw.edu} \\
  \And
  Chirag Shah\\
  University of Washington\\
  Seattle, WA, USA \\
  \texttt{chirags@uw.edu} \\
}

\begin{document}

\maketitle

\begin{abstract}
  Often machine learning models tend to automatically learn associations present in the training data without questioning their validity or appropriateness. This undesirable property is the root cause of the manifestation of \emph{spurious correlations}, which render models unreliable and prone to failure in the presence of distribution shifts. Research shows that most methods attempting to remedy spurious correlations are only effective for a model's known spurious associations. Current spurious correlation detection algorithms either rely on extensive human annotations or are too restrictive in their formulation. Moreover, they rely on strict definitions of visual artifacts that may not apply to data produced by generative models, as they are known to hallucinate contents that do not conform to standard specifications. In this work, we introduce a general-purpose method that efficiently detects potential spurious correlations, and requires significantly less human interference in comparison to the prior art. Additionally, the proposed method provides intuitive explanations while eliminating the need for pixel-level annotations.
  We demonstrate the proposed method's tolerance to the peculiarity of AI-generated images, which is a considerably challenging task, one where most of the existing methods fall short. Consequently, our method is also suitable for detecting spurious correlations that may propagate to downstream applications originating from generative models.
\end{abstract}

\section{Introduction}

Spurious correlations in machine learning (ML) models arise from uncontrolled confounding biases present in data collection \cite{arjovsky2019invariant}. They render ML models vulnerable to distribution shifts and lead to poor generalization  \cite{taori2020measuring}. Several fundamental flaws of ML models can be traced back to spurious correlations, such as bias and poor robustness. Therefore, to make these models reliable in deployment, a necessary first step is to identify spurious correlations accurately.

With the recent surge in the usage of generative models such as DALL-E 2 \cite{ramesh2021zero}, CLIP \cite{radford2021learning}, and others, AI-generated data is increasing its footprint on the internet. Eventually, we can expect training data for ML models to significantly comprise such content as they are already being used for performing data augmentation \cite{ijcai2023p659,akrout2023diffusion,trabucco2023effective}. Hence, it is paramount to devise a method that supports detecting spurious correlations in downstream models that were trained on AI-generated data. Problematic content generated by a single foundation model could affect numerous applications, leading to a considerable propagation of undesirable traits. 

Vision-language models were shown to generate images with bizarre visuals, especially when presented with prompts that entail compositional reasoning and complex spatial relationships between objects \cite{thrush2022winoground,ramesh2022hierarchical}. When dealing with such images, the existing spurious correlation detection algorithms may prove to be ineffective. This outcome is expected due to the reliance of these methods on segmentation \cite{plumb2022finding}, or strict definitions of visual artifacts \cite{singla2022salient}. 

Notably, a method that can automatically distinguish between causal and spurious patterns without any human intervention is the most desirable solution. However, this would require a causal machine learning model that is trustworthy and without flaws -- and such models do not yet exist for image classification. 
All of the existing methods for detecting spurious patterns rely on some form of human knowledge, either directly or through algorithmic encoding, and this dependence may not be solved without major breakthroughs in causal machine learning. Until then, the primary goal is to reduce the manual effort involved, along with a few more desirable properties that we address through the proposed method.

This paper addresses many of the existing shortcomings. Specifically, the contributions of the proposed method are:

\begin{itemize}
    \item Automatically detects potential spurious correlations.
    \item Provides intuitive visual explanations which help in confirming valid spurious correlations.  
    \item Automatically flags similar instances of an identified spurious correlation.%, significantly reducing manual effort. %the need for human intervention.
    \item Compatibility with AI-generated images.
    \item Eliminates the need for object segmentation and pixel-wise annotations.
\end{itemize}

\section{Related Work}

% Despite the importance of detecting spurious correlations automatically, not much work has been done to address this issue until recently. 
In recent years, a few papers have been published in the literature that study spurious correlations in text \cite{wang2021robustness, wang2021identifying, srivastava2020robustness} and a few on vision models \cite{singla2022salient, plumb2022finding, adebayo2022post, ming2022impact, wong2021leveraging}. However, none of these methods consider the complications that arise when working with generative models. Furthermore, the existing spurious correlation detection methods in vision either rely on extensive human interference or pixel-level annotations.

While explainability techniques could be used to spot spurious correlations in theory, they are not sufficiently suitable for this task as they typically generate local explanations for each sample \cite{chattopadhay2018grad, simonyan2013deep, selvaraju2017grad, ross2017right, singh2018hierarchical, dhurandhar2018explanations, goyal2019counterfactual, wang2020score}. They do not aggregate information by detecting repeated occurrences of spurious patterns, and making manual observations by probing each explanation is infeasible. Moreover, studies have shown that these methods may not always be reliable \cite{adebayo2022post, adebayo2020debugging}. To address these challenges, a few dedicated techniques for identifying spurious correlations have been proposed which build on top of explainability methods \cite{singla2022salient, hagos2022identifying, plumb2022finding}.

In \cite{plumb2022finding}, the authors propose a method that identifies spurious patterns on the basis of sensitivity to changes from counterfactual versions of the original instance, and have a human label the identified patterns as spurious or valid. In \cite{singla2022salient}, the authors show that obtaining human annotations for strategically procured representative images can generalize well for unseen images. Yet, this approach still required a significant amount of manual intervention that warranted crowd-sourcing. In \cite{hagos2022identifying}, the authors proposed using explanation-based learning to identify spurious correlations, however, this study was conducted on toy datasets and it is unclear how well it would generalize.

\begin{figure}[h]
  \centering
  \includegraphics[width=1\textwidth]{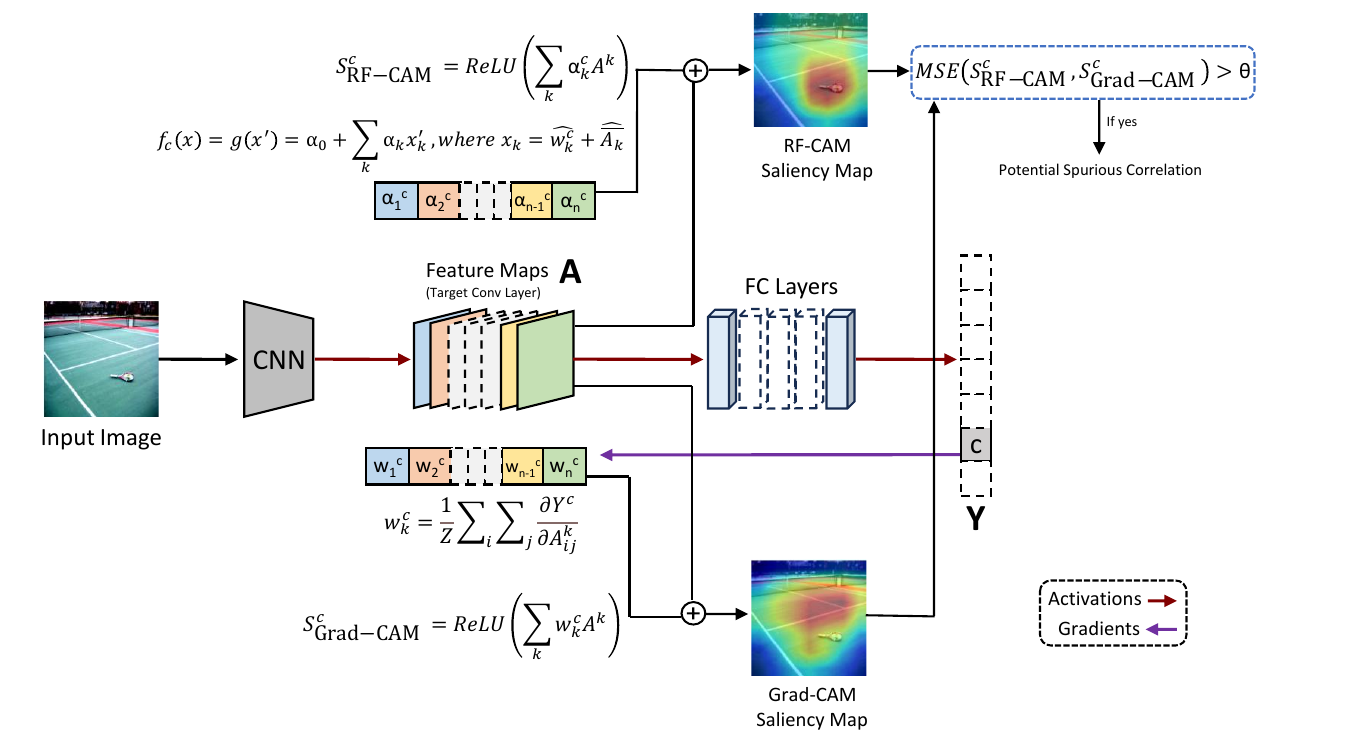}
  \caption{Overview of the proposed method for detecting \emph{spurious correlations} via robust features.}
  \label{fig:overview}
\end{figure}

\section{Generic Framework for Spurious Correlation Detection}

\subsection{Core and Spurious Attributes}
\label{definitionsSSC}
In vision tasks, the definition of spurious attributes is a bit vague, but previous methods have established a usable interpretation \cite{singla2022salient, plumb2022finding}. Similar to the definition presented in \cite{singla2022salient}, we consider 
\begin{itemize}
    \item visual features always expected to be part of the object definition as \emph{core attributes}
    \item features that are likely to co-occur, but not a part of the object, as \emph{spurious attributes}.
    
\end{itemize}

However, we extend the definition of \emph{core attributes} to include features that are perceivably a part of the object but may not always be expected to be present. This modification enables us to account for the hallucinations or imperfections of generative models. For instance, if a generative model depicts a gymnast as a person with three legs, all of the limbs should be considered \emph{core attributes} even though this deviates from physical reality -- as it reflects the generative model's perception of reality. As a result, the flexibility of this formulation allows us to understand the generative model's worldview and examine its spurious associations.

\subsection{Approach Overview}

We propose building local surrogate models (LSMs), one for each image class, trained to learn robust features specific to its respective class. As these robust features are representative of core attributes, disagreement between the image classifier and the LSM of the predicted class can reveal spurious associations.
In order to accommodate deviations from conventional definitions, we utilize prototypical explanations that convey spurious behavior. This makes the framework sufficiently flexible to support any data distribution, even when the contents of the images are not photorealistic, as we only need to point out the spurious concepts through heatmaps without requiring to define them. 
We operationalize this approach by leveraging feature activations in the final convolutional layer, gradients, and the explanations from LSMs designed to learn robust features and the behavior of the image classifier for each class.

\subsection{Methodology}
Here, we present the details of our method in logical parts, and then tie it all together in \ref{subsec:FlowOfOps}.

\subsubsection{Spurious Correlation Identification}
\label{subsec:SCidentification}

As per the definitions in \ref{definitionsSSC}, each input image $x \epsilon X$ has core attributes $c \epsilon C$ and spurious attributes $c^\prime \epsilon$ $U$ $\setminus$ $C$, where $U$ is the universal set of all features. Consider an image classifier $f(x)$ : $X \rightarrow Y$, where each $x \epsilon X$ has to be assigned a label $y \epsilon Y$. Our goal is to identify if any $c^\prime$ attributes were activated when $x$ is correctly classified, and flag such instances for potential spurious correlations.

\subsubsection{Local Surrogate Models (LSMs)}
\label{subsec:localSurr}
Without groundtruths, either through pixel-level annotations or human review, detecting if a visual attribute is spurious is non-trivial. Therefore, we build an LSM for each class that bases its decisions on robust features, as they tend to represent core attributes. 
Global models that leverage robust features were shown to significantly minimize reliance on spurious attributes \cite{wong2021leveraging,singla2021understanding}, however, they demonstrate lower clean accuracy. In contrast, our approach leaves the original model unaltered and builds LSMs that help us detect and visualize spurious correlations.

For an image classifier $f(x)$ : $X \rightarrow Y$, where each $x \epsilon X$ is assigned a label $y \epsilon Y$, we build an LSM for each class $c$, $f_c(\phi)$ : $\Phi \rightarrow Y^{\prime}$, where each $\phi \epsilon \Phi$ is representative of an image $x$ in a transformed latent space $\Phi$, and $y\prime \epsilon Y^{\prime} ;  y\prime \in \{0,1\}$ represents if the original model has misclassified. Each class $c$ will have an LSM $f_c$, which is trained and evaluated only on instances belonging to $c$. 

Each image $x$ is appropriately represented by its corresponding $\phi$, as it is the element-wise sum of the global-average-pooled gradients of original model's prediction w.r.t. activation maps $A^k$

\begin{equation}
w_k^c=\frac{1}{Z} \sum_i \sum_j \frac{\partial Y^c}{\partial A_{i j}^k}
\label{eq:wkc}
\end{equation}

($Z$ = number of pixels in $A^k$ with dimensions $i*j$)

and the mean of activation map $\bar{A}^k$. Both $w_k^c$ and $\bar{A}^k$ are normalized to ensure equal weightage.

\begin{equation}
\phi_k^c =  \widehat{w_k^c} + \widehat{\bar{A}^k}    
\label{eq:akc}
\end{equation}

Intuitively, $\widehat{\bar{A}^k}$ represents any visual attributes present in an image, and $\widehat{w_k^c}$ represents the visual attributes contributing to the decision. We want our LSM $f_c(\phi)$ to learn patterns from both of these entities in relation to misclassification, and therefore we perform an element-wise sum. 
While we use XGBoost \cite{chen2016xgboost} as the model choice for LSMs, any ML method could be used.

\subsubsection{Computing SHAP values}
\label{subsec:computeShap}
When an input image is classified as class $c$, we employ the LSM for that class, $f_c(\phi)$, for generating SHAP \cite{lundberg2017unified} values corresponding to visual attributes in the image. SHAP assigns each feature an importance value for a particular prediction

\begin{equation}
f_c(\phi) = g(\phi^\prime) = \alpha_0 + \sum_k \alpha_k \phi_k^\prime    
\label{eq:fc}
\end{equation}

The explanation model $g(\phi^\prime)$ approximates the surrogate model $f_c(\phi)$ and computes contributions $\alpha_k$ by every feature, and these values are used to generate the robust features saliency maps.

\subsubsection{Robust Features Saliency Map (RF-CAM)}
\label{subsec:RFCAM}
To produce robust features class activation maps (RF-CAM), a weighted combination is performed on SHAP values computed in equation \ref{eq:fc} and class activation maps, followed by a ReLU \cite{agarap2018deep}:

\begin{equation}
S_{\text {RF-CAM }}^c=\operatorname{ReLU} \left(\sum_k \alpha_k^c A^k\right)
\label{eq:rfcam}
\end{equation}
RF-CAM saliency maps highlight robust features characteristic to the predicted class, which tend to be core attributes. In contrast, explainability methods such as Grad-CAM \cite{selvaraju2017grad}, highlight visual features that triggered the prediction by leveraging $w_k^c$ computed in equation \ref{eq:wkc}, and observing the difference between these saliency maps reveals information regarding potential spurious attributes. 

\begin{equation}
S_{\text {Grad-CAM }}^c=\operatorname{ReLU} \left(\sum_k w_k^c A^k\right)    
\label{eq:gradcam}
\end{equation}

ReLU \cite{agarap2018deep} is used on the weighted combination to preserve positively associated robust features toward a particular class. These saliency maps will be the size of the featuremaps in the final convolution layer, and they are upscaled to the input image size, and overlayed as the final step.

\subsubsection{Spurious Correlations from Saliency Maps}
\label{subsec:SCfromSM}
Consider that the RF-CAM accurately highlights core attributes of the predicted class in an image, and an explainability map, such as Grad-CAM, accurately highlights the attributes that triggered the prediction. In the presence of a spurious correlation, a correctly classified image will have these two saliency maps that differ beyond a threshold $\theta$. In an ideal scenario when there is no spurious correlation, these saliency maps should look very similar. We can measure the dissimilarity using Mean Squared Error: 
$MSE(S_{\text {RF-CAM }}^c, S_{\text {Grad-CAM }}^c) > \theta \label{eq:mse}$ The appropriate threshold $\theta$ limit will vary depending on several factors such as model, dataset, and surrogates.

\subsubsection{Flow of Operations}
\label{subsec:FlowOfOps}

The overview of the proposed method is presented in Figure. \ref{fig:overview}.
As a first step, using the original image classifier on each sample, we compute the predicted label, feature maps $A^k$, and the gradients $w_k^c$ for the predicted class $y_c$. All of these entities are required for training LSMs.

Using the labels predicted by the image classifier and comparing them with true labels, a new misclassification label is generated for all samples in both training and test sets. Next, for each class $c$, a LSM $f_c$ is trained using the misclassification labels on the train split and evaluated using the test split. The input for these surrogate models is the combination of unit gradients and mean activations, given by $\theta$ (discussed in detail in \ref{subsec:localSurr}). SHAP values generated on these local surrogates represent importance values for robust features (refer \ref{subsec:computeShap}), and these values are used to produce robust feature saliency maps (RF-CAM) (refer \ref{subsec:RFCAM}). 

Subsequently, comparing RF-CAM and Grad-CAM maps using mean square error, potential spurious correlations are identified. These findings on potential spurious correlations are then presented to the model developer, as human judgment and domain knowledge are vital for the confirmation of spurious behavior. For a confirmed spurious correlation, the most activating neural feature can be utilized to identify other instances of this spurious correlation. In \cite{singla2022salient}, the authors show that neural feature annotations generalize very well to many images.

\section{Datasets}

We evaluate our method on three different datasets.
The first one is the widely-used ImageNet \cite{deng2009imagenet}, which was also used in other studies exploring spurious correlations \cite{singla2022salient,wong2021leveraging,singla2021understanding}. 
As our method supports content produced by generative models -- an improvement over extant work, we would require synthetic datasets generated by an AI model to demonstrate this property. Therefore, we create two datasets that are AI-generated, using Stable Diffusion 2.1 \cite{rombach2022high}. 

\emph{\textbf{GenAI Images Dataset 1}}: 9,750 images are generated for 75 classes of \emph{MS-COCO} \cite{lin2014microsoft} using dataset captions. All classes are balanced, with 100 training images and 30 test images each.

\emph{\textbf{GenAI Images Dataset 2}}: 
52,000 images are generated for 40 classes of \emph{ImageNet People SubTree} \cite{yang2020towards} using Vit-GPT2 \cite{dosovitskiy2020image} captions. Each class has 1,000 training and 300 testing images.

Further details of GenAI images datasets are discussed in Appendix (refer \ref{subsec:datasets}).

\section{Experiments and Results}

\begin{table}[h]
\centering
\caption{Accuracies of Image Classifiers (IC) and averaged acc. of Local Surrogate Models (LSM). \label{perfTable}}
\begin{tabular}{@{}lllll@{}}
\toprule
\textbf{Classifier} & \textbf{Dataset} & \textbf{Classes} & \textbf{IC Acc} & \textbf{LSM Avg. Acc} \\ \midrule
ResNet50            & ImageNet         & 1,000             & 74.54        & 99.93            \\
ResNet50            & GenAI Images 1   & 75               & 88.56        & 96.58            \\
ResNet50            & GenAI Images 2   & 40               & 83.19        & 95.28            \\ \bottomrule
\end{tabular}
\end{table}

\begin{figure}[h]
  \centering
  \includegraphics[width=\linewidth]{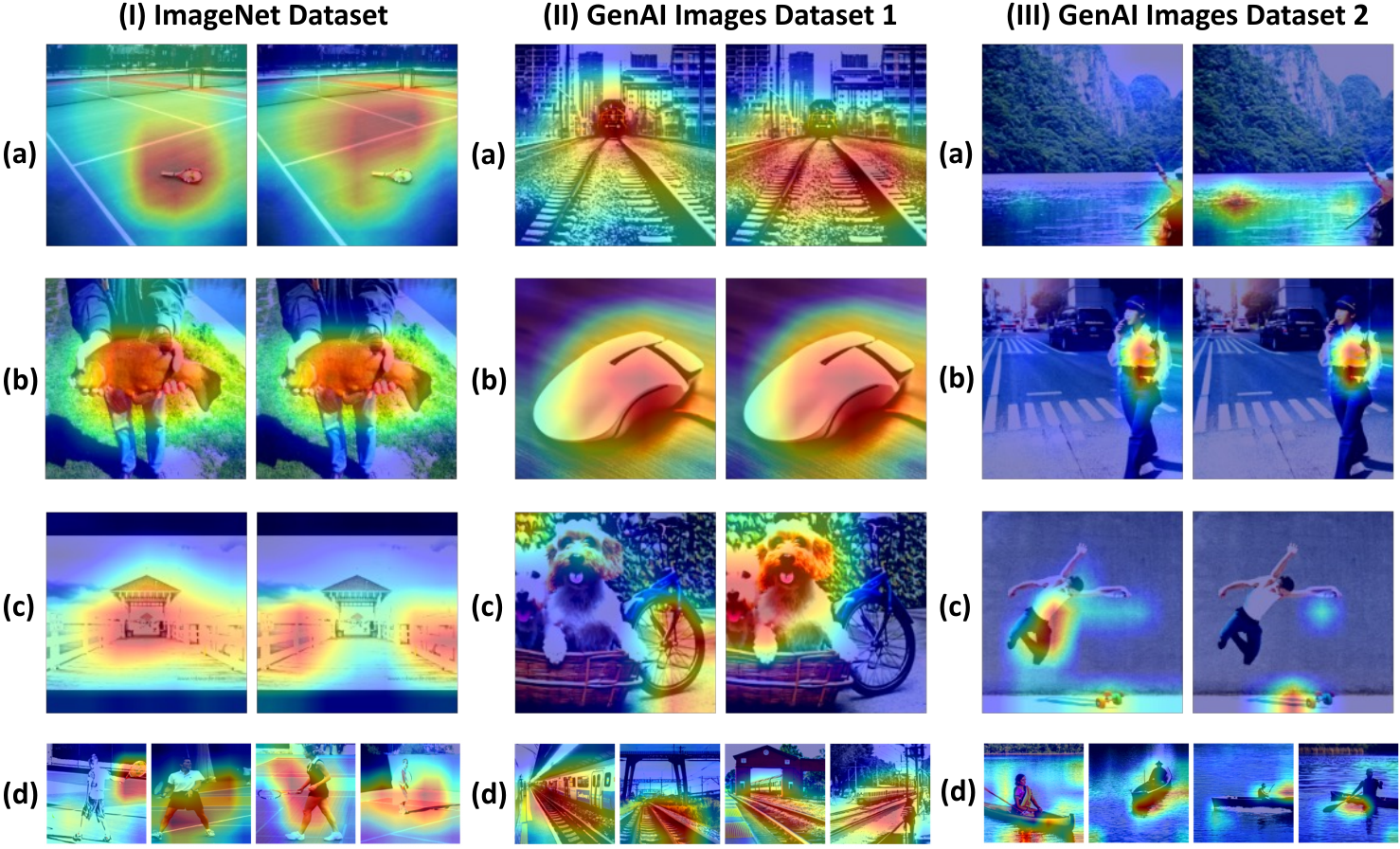}
  \caption{RF-CAM and Grad-CAM (left and right respectively, in each pair for [a-b]). Correctly classified instances based on (a) Spurious Attributes (b) Core Attributes; (c) Misclassified instances (d) Grad-CAM of instances activating spurious feature (I) 1890 for class \emph{racquet} (II) 936 for class \emph{train} (III) 1017 for class \emph{boatperson}.}
  \label{fig:imgsRes}
\end{figure}

\begin{figure}[h]
  \centering
  \includegraphics[width=0.8\linewidth]{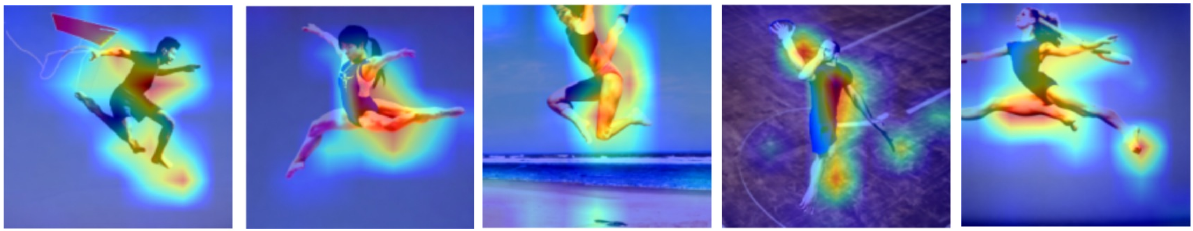}
  \caption{RF-CAM highlights deformities as representative and robust attributes for class \emph{Gymnast}}
  \label{fig:genImgFlaws}
\end{figure}

We present the findings of our method on three datasets with significantly different data distributions. While ImageNet is representative of real-world images, the GenAI datasets are suitable for testing the compatibility of our method with AI-generated images. \emph{GenAI Images 1} consists of relatively more photorealistic images, while images in \emph{GenAI Images 2} are characterized by peculiar imperfections since Stable Diffusion does not perform very well for human images.

We choose pretrained ResNet50 \cite{he2016deep} as the image classifier, but any CNN-based vision model is supported. Default weights were used on ImageNet, and only the last FC layer was retrained to match the number of classes in the other two datasets.
We tabulate the performance of the image classifier and the average accuracy of the LSMs in Table. \ref{perfTable}. The high LSMs average accuracies indicate that the robust features learned by them are reliable.

In summary, in all three datasets, the proposed algorithm successfully identified potential spurious correlations, generated intuitive explanations to help aid decisions, and retrieved similar instances for identified spurious correlations. Roughly, 35\% to 40\% potential spurious correlations were detected in each test set. Notably, many of these instances are repetitions belonging to a few spurious associations. When the model developer confirms a detected potential spurious correlation, several similar spurious instances can be remedied automatically. Therefore, the manual labor involved is significantly reduced, and notably lower than other existing methods. Similar to \cite{singla2022salient}, Human Intelligence Tasks (HITs) would be required to compute the conversion ratio of potential spurious correlations to confirmed ones, and is a viable task for future work.

In all experiments, the dissimilarity between RF-CAM and Grad-CAM heatmaps (section \ref{eq:mse}) was calculated on pixels with intensity higher than 0.78, which roughly correspond to the red regions in the heatmaps. This prevents noise from significantly affecting the dissimilarity score. Additionally, all pixels with more than 0.78 intensity are rounded to 1 while computing the MSE, in order to increase tolerance to minor variations. The MSE threshold ($\theta$) is set to 15.

In the first row of Figure. \ref{fig:imgsRes}, we observe instances where the classifier has based its predictions on spurious attributes. The first heatmap in each pair is RF-CAM, which highlights core features and the second heatmap highlights the features that were actually used by the classifier, and differences between these uncover spurious attributes. For instance, in (Ia), we notice that the classifier relies on the tennis court background rather than the racquet itself. By using the most activated neural feature, we obtain other instances of this spurious association between the class \emph{racquet} and tennis court, and the top four retrieved images are shown in (Id). Similarly, in (IIa), we observe that the classifier relies on railway tracks to identify trains, and other instances of this spurious association are shown in (IId), some of which have no train visible at all. In (IIIa), we notice that only water background is used for identifying boatperson and other instances of this can be seen in (IIId). In the second row, we observe that when there are no spurious associations present, RF-CAM and Grad-CAM heatmaps look almost identical, as expected.

In images with multiple objects, there can be several valid labels. Most classification tasks, however, only designate one of these as the `groundtruth' label. When an image classifier identifies a valid label that is different from the groundtruth, RF-CAM can be utilized to emphasize the key features of the groundtruth class. This is particularly helpful for debugging models, as it shows which areas the model should have focused on for a correct prediction. For example, in (Ic), RF-CAM highlights the area the classifier should have considered to classify it as \emph{`pier'}, while Grad-CAM illustrates the area that led to its incorrect classification as \emph{`worm fence'}.

In Figure. \ref{fig:genImgFlaws}, we notice that multiple RF-CAM heatmaps show malformed limbs as robust features representative of a gymnast. This demonstrates that the proposed method can be useful for exploring and uncovering spurious associations projected by the generative model as well.

Additional details on results for each dataset are discussed in Appendix \ref{subsec:FurtherRes}. 

\section{Limitations}
There is no one solution that fits all requirements for spurious correlation detection. The proposed method, for instance, will perform better when there are sufficient misclassifications in the dataset that were caused by spurious associations. Noticeably, this is commonly observed in most datasets with unconstrained real-world settings, as they would have long tails of data distribution and memorization becomes unlikely. Conversely, if a spurious correlation is so prevalent in the dataset and without counter-examples, the LSMs might pick it up as a robust feature. In such cases, where the dataset is not sufficiently diverse, evaluating on a different dataset built for the same purpose would be helpful in overcoming the shortcomings.

In scenarios where an image contains multiple objects, the proposed method may flag for potential spurious correlations when the predicted label deviates from the groundtruth label, as illustrated in Figure \ref{fig:imgsRes} (I-IIIc). However, this issue arises from the restrictive formulation of classification tasks where only a single groundtruth label is considered even when multiple objects are present in the image.

\section{Conclusion}
In this paper, we proposed a method that effectively detects spurious correlations while providing several advantages over the existing methods, such as: (1) not requiring pixel-level annotations; (2) unconstrained by restrictive object definitions; (3) intuitive visual and comparative explanations; (4) automatic flagging of similar spurious correlation instances; and (5) compatibility with AI-generated images. Furthermore, we discussed the importance of this line of work in the currently evolving landscape of visual data, driven by a surge in the adoption of generative models.

\section*{Acknowledgements}
This work was supported in part by the cloud computing credits made available by the University of Washington eScience Institute in partnership with Microsoft Azure.

\bibliographystyle{plainnat}
\bibliography{references}

\newpage

\appendix

\section{Appendix}
% \section*{Supplementary Material}
% \renewcommand{\thesection}{\Alph{section}}

\subsection{Further Dataset Details}
\label{subsec:datasets}

\subsubsection{GenAI Images Dataset 1}
\label{subsec:AIgenDataset1}
To demonstrate that the proposed method works well with AI-generated images, we create a new dataset using Stable Diffusion 2.1. As this is the first work of its kind, a standard dataset of text prompts for generating images does not yet exist. To circumvent this issue, we leverage an image captioning dataset -- MS-COCO \cite{lin2014microsoft}, and generate the dataset by using the captions as the input prompts to Stable Diffusion. We noticed that, in many cases, the generated image and the original image which share the same caption/prompt are not very representative of each other. Also, each image in MS-COCO has 5 captions, all of which could yield very different results. 

There are 80 object categories in the MS-COCO dataset, and each image has pixel-wise annotations for these objects, if present. As our goal is to generate a classification dataset, we use object categories as class labels for images that contain these objects. Note that an image can contain multiple objects, and in those cases, we randomly pick one of the object labels as the class label. This results in a labeling scheme very similar to ImageNet, as it also contains images with multiple objects and a single class label. This property is very representative of the real world in unconstrained settings.

It is imperative to ensure that the object corresponding to the class label is present in the generated image. To this effect, we filter and select the captions with the class label explicitly mentioned in them, forcing the generative model to produce visual features corresponding to the class label. Applying this constraint resulted in the removal of 4 categories, which are \emph{hair drier}, \emph{sport ball}, \emph{handbag}, and \emph{toaster}, as they have very few captions explicitly consisting of them as words. Additionally, we also removed the \emph{person} category, as most images contain people in the background, and this adds a significant amount of perplexity to the task. For the remaining 75 categories, we generated 130 images each -- 100 to be included in the training set and 30 in the test set. This results in a total of 9750 images, with 7500 in the train split and 2250 in the test split. Most of the generated images in this dataset were photorealistic, indicating that the image captions are suitable inputs to the generative model.

\subsubsection{GenAI Images Dataset 2}
\label{subsec:AIgenDataset2}
Stable Diffusion 2.1 performs impressively for non-human images and achieves remarkable photorealism, but it seems to fall short when humans are involved as the main subject of the image. For a thorough evaluation, we require a synthetic dataset that is characterized by peculiar imperfections of AI-generated images. Therefore, we create a synthetic version of ImageNet people subtree.

\emph{ImageNet People SubTree \cite{yang2020towards}} is a subset of ImageNet containing 2,832 people categories. Although, most of these categories are unusable as only 139 of them are considered safe and imageable \cite{yang2020towards}. From these 139 classes, we use ones that are either a profession or an occupation. Additionally, categories that share a lowest common hypernym and have a broader yet specific definition of an occupation were merged together. For instance, \emph{captain}, \emph{chief\_of\_staff}, \emph{general}, \emph{major}, \emph{Navy\_SEAL}, \emph{military\_personnel} were all mapped to \emph{military\_officer}, as some of these classes are an abstraction of others. Finally, we get 40 classes that are used in this study. %cite IJCAI paper? Similarity checks?

The synthetic version of ImageNet people subtree is created by first generating captions for each image using Vit-GPT2 \cite{dosovitskiy2020image,radford2019language}. However, most of these captions are not granular enough to describe details about the person and simply use pronouns. In order to overcome this issue, we replace all pronouns with the person class label corresponding to the image, and this step results in rich text prompts that yield images representative of the person category. Finally, these captions are fed as text prompts to generate images using Stable Diffusion. For each of the 40 classes, we generate 1,000 training images and 300 test images, leading to a training split of 40,000 images and a test split of 12,000 images.

\subsection{Difficulties in handling AI-generated images}

Generative models are known to produce images that are incomplete, confusing, and sometimes incomprehensible. This is especially noticeable when the contents of the image are complex, require compositional reasoning, or relational constraints.

\begin{figure}[h]
  \centering
  \includegraphics[width=0.6\linewidth]{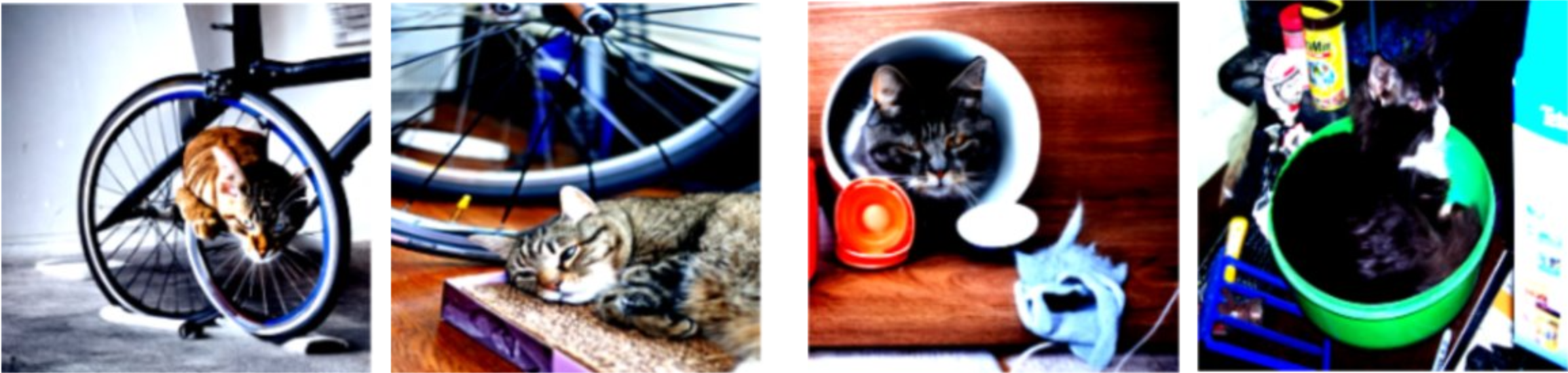}
  \caption{Flaws in AI-generated images. The first image in each pair is synthetic, and the second is a real image sharing the same caption.}
  \label{fig:compositional}
\end{figure}

In Figure. \ref{fig:compositional}, the first pair of images share the caption: \emph{`A tabby cat laying on a cat scratcher in front of a bicycle wheel'}, and Stable Diffusion generated a cat coming out of a bizarre-looking bicycle wheel. Second pair share the caption \emph{`A cat resting inside a bowl next to different household items'}, and the generated image has a cat's head placed in a bowl, and none of the other objects are recognizable. These synthetic images are from GenAI dataset 1 [\ref{subsec:AIgenDataset1}].

\begin{figure}[h]
  \centering
  \includegraphics[width=0.8\linewidth]{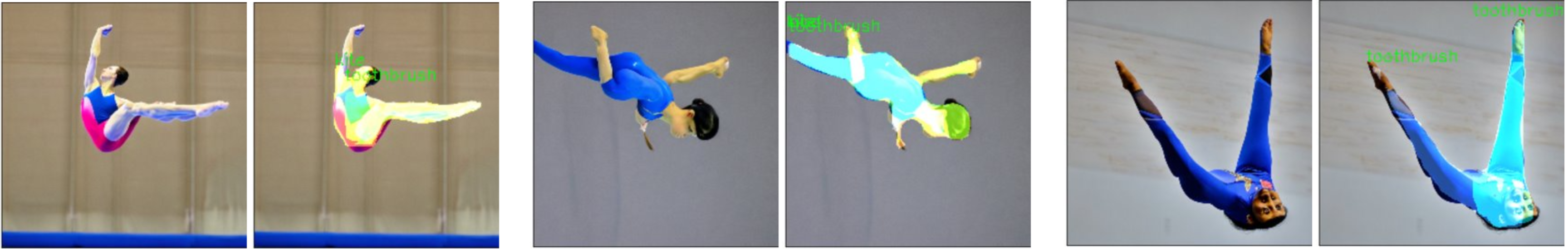}
  \caption{Segmentation Failure. Prompt: \emph{`a gymnast in a blue shirt is on a blue and white board'}. Instead of \emph{person}, detected as \emph{kite} \& \emph{toothbrush}.}
  \label{fig:segFail_gymnast}
\end{figure}

Such peculiar flaws in AI-generated images pose a challenge to existing spurious correlation detection algorithms, as they cause segmentation methods to fail. For instance, consider the images in Figure. \ref{fig:segFail_gymnast}, in all three images a gymnast was not detected as a person and instead as a \emph{kite} or \emph{toothbrush}. These synthetic images are from GenAI dataset 2 [\ref{subsec:AIgenDataset2}].

\subsection{Further Result Details}
\label{subsec:FurtherRes}

\emph{\textbf{ImageNet: }}
The ImageNet \cite{deng2009imagenet} dataset is highly representative of images taken in the wild, and consists of strong signals that co-occur together in the real world, making it ideal for testing spurious correlation detection algorithms. Prior works addressing spurious correlations in image classification have conducted experiments on ImageNet \cite{singla2022salient,wong2021leveraging,singla2021understanding}. On a test set of size 50,000, our method detected 18,062 (36.12\%) instances of potential spurious correlations.

\emph{\textbf{GenAI Images Dataset 1: }}
In order to demonstrate the compatibility of our method with AI-generated images, we evaluate it on this dataset generated from MS-COCO image captions using Stable Diffusion. Details of this dataset are discussed in Section \ref{subsec:AIgenDataset1}. Stable Diffusion was able to generate photorealistic images based on COCO captions with few exceptions. Our method handled this dataset well, with a performance similar to ImageNet. Out of 2,250 images in the test set, 923 (41.02\%) potential spurious correlations were detected.

As most of the images in this dataset are photorealistic, it is unsurprising that the method is able to perform equally well. However, it should be noted that traditional segmentation algorithms might still fail at higher rates for AI-generated images, even if they appear to be mostly photorealistic, as they are still out-of-distribution samples to the segmentation model.

\emph{\textbf{GenAI Images Dataset 2: }}
The tolerance toward the peculiarity of AI-generated images of the proposed algorithm is better demonstrated on this dataset, as Stable Diffusion does not perform very well for human images. Unlike GenAI images dataset 1 [\ref{subsec:AIgenDataset1}], this dataset is not photorealistic and contains evident imperfections. Despite these flaws, the proposed algorithm has performed similarly to other datasets, demonstrating its tolerance and compatibility with AI-generated images. Out of 12,000 test images, 4,103 (34.19\%) were detected to have potential spurious correlations.

We note that defining core attributes for the classes of ImageNet person subtree is a complex task and without a clear solution, as many of these classes have an imageable score of less than 1 \cite{yang2020towards}. However, the purpose of this experiment is to test the ability of the proposed method in handling the imperfections of AI-generated images, and these flaws appear to be more prevalent in human-related content, making this dataset a suitable option.

\subsection{Observations and Opportunities}

Arguably, one of the most crucial problems of this decade that has to be addressed by the machine learning community, in light of the recent advances in generative models, is to upgrade current information systems to effectively process AI-generated content. Generative AI is now used in the creation of art \cite{srinivasan2021biases,wang2023reprompt, oppenlaender2022taxonomy}, music \cite{plut2020generative,berg2023case}, writing \cite{muller2022genaichi,chung2022talebrush}, and many more applications.

Notably, almost every modern information access system, either directly or indirectly, relies on machine learning. These include infotainment services like music recommendations, and applications with more serious implications such as job search and medical advice. It is paramount to ensure that these ML services are robust and reliable. A major step toward this goal is verifying that they are free from spurious correlations. Apart from compromising performance in deployment, researchers have shown that spurious correlations have societal consequences by causing the amplification and propagation of bias \cite{struppek2022biased,hendricks2018women,zhao2017men,bender2021dangers}.

Information retrieval systems, especially the ones that leverage neural ranking and neural recommendation, will be significantly affected by this new incursion of AI-generated content that is becoming ubiquitous on the web. The National Institute of Standards and Technology (NIST) has stressed the importance of identifying limitations and scenarios where machine learning models could fail \cite{258151}. This goal cannot be realized unless advances are made in detecting spurious correlations.

%to uncover spurious correlations. 
% the model was fitted very well to the dataset and has very few misclassifications,

%Another fundamental approach that shows promise in addressing spurious correlations is invariant risk minimization \cite{arjovsky2019invariant}, a learning paradigm that promotes the estimation of invariant correlations across multiple training distributions. If there are multiple datasets, all of which having their own set of biases, learning patterns that are common across all these distributions while ignoring the spurious ways in which they differ, will result in robust models with relatively few spurious correlations. 

Once spurious correlations are accurately identified, there are several ways to remedy them. Model retraining is the most common approach, which is achieved by retraining the model on an augmented dataset -- either by removing harmful samples or by introducing neutralizing samples. Procuring new samples that act as counterbalances can be achieved through different methods, and techniques such as counterfactual data generation, GANs, VAEs, simulation engines, and more recently generative models have been explored \cite{ijcai2023p659,plumb2022finding,jaipuria2020deflating,yucer2020exploring,hu2019exploring,wu2019data}. 

Another future direction, one with potentially tremendous impact, is detecting and quantifying spurious correlations in foundation models. These models are adapted for various tasks, and any bias originating from them would have far-reaching effects in downstream applications. Compared to discriminative models, uncovering spurious associations of generative models appears to be more challenging. One approach to solving this problem could be through a proxy model, where a downstream model trained on the content generated by a foundation model is examined to understand the spurious behaviors of the main model. Addressing the problem of spurious correlations at the source, i.e., generative models, could translate into a large impact downstream and eventually lead to cleaner data in information systems and the web. 
%%%%%%%%%%%%%%%%%%%%%%%%%%%%%%%%%%%%%%%%%%%%%%%%%%%%%%%%%%%%

\end{document}